\documentclass[conference]{IEEEtran}
\IEEEoverridecommandlockouts
\usepackage{cite}
\usepackage{amsmath,amssymb,amsfonts}
\usepackage{algorithmic}
\usepackage{graphicx}
\usepackage{textcomp}
\usepackage{xcolor}
\usepackage{subfig}
\usepackage{booktabs}
\usepackage{import}
\usepackage{color}
\usepackage[citebordercolor=blue]{hyperref}
\usepackage[capitalise]{cleveref}
\usepackage{multirow}
\usepackage{array}
\usepackage[font=footnotesize]{caption}
\usepackage[inline]{enumitem}
\usepackage{makecell}
\usepackage{siunitx}
\usepackage{caption}
\begin{document}
%
\title{Detecting Small Space Objects for Space Situational Awareness with Collaborative Edge AI}
%
%
%

\author{\IEEEauthorblockN{Xingyu~Qu~and~Peng~Hu}
\IEEEauthorblockA{
Advanced Network and Embedded Systems Lab\\
Department of Electrical and Computer Engineering, University of Manitoba, Canada R3T 2N2\\
  {que1@myumanitoba.ca, peng.hu@umanitoba.ca} \\ }
\thanks{We acknowledge the support provided by the Natural Sciences and Engineering Research Council of Canada (NSERC), [funding reference number RGPIN-2022-03364], and Research Manitoba.}} 

%
%

\markboth{Journal of \LaTeX\ Class Files,~Vol.~14, No.~8, August~2015}%
{Shell \MakeLowercase{\textit{et al.}}: Bare Demo of IEEEtran.cls for IEEE Journals}
%



\maketitle

\begin{abstract}
With the rapid growth of space activities, near-Earth space is becoming increasingly congested, leading to a higher risk of spacecraft collisions and space debris generation. Recent close-proximity encounters and collision incidents necessitate efficient and cost-effective onboard systems for space object detection (SOD) and collision avoidance. While deep learning (DL) enabled Edge AI using low-cost vision sensors on single or multiple satellites has shown promises for SOD, significant challenges remain, including limited datasets, small SOD, and the lack of comprehensive performance assessments. This paper presents a collaborative Edge AI framework for multiview SOD and systematically benchmarks DL models and fusion strategies. Results show that higher image resolution significantly improves small SOD. For multiple views, image-level fusion reduces the combined communication and inference latency by 30.13\% and 32.80\% on average compared with feature-level and attention-based fusion across the evaluated inter-satellite link (ISL) bandwidth. For larger view sets, feature-level and attention-based fusion achieve better detection accuracy at the cost of analysis in the sensing-communication pipeline, revealing that multiview collaboration introduces minimal latency overhead under ISL constraints when reception capacity is not exceeded. Our work provides valuable design insights for future collaborative satellite network design, supporting the emerging space situational awareness and space traffic management requirements. 

\end{abstract}



%
\IEEEpeerreviewmaketitle

\section{Introduction}
As space activity intensifies, near-Earth space is becoming increasingly crowded with an unprecedented number of satellites and other space assets. The risk of Kessler syndrome underscores the urgent need to deploy technologies that prevent collisions and mitigate the generation of space debris. Failure to act could jeopardize future space endeavors, including the development of a circular space economy, space data centers, and interplanetary missions. One promising approach is vision-based sensing, where space-grade vision sensors integrated with deep learning (DL)-enabled edge AI inference are deployed on satellites to detect and track space objects, thereby improving space situational awareness (SSA). While this approach offers several advantages, such as energy efficiency, low cost, and straightforward deployment, maintaining high detection accuracy remains challenging due to the wide variation in object sizes, especially for space objects, whose limited visual signatures make reliable detection difficult. The scarcity of publicly available datasets for this task further exacerbates the problem, hindering the training, validation, and benchmarking of advanced DL models. Consequently, reliable detection of small space objects remains a key challenge for vision-based SSA systems.

Current solutions are based on the YOLO-based models\cite{ai_driven_collab, Wenxuan_ccece_2025, sensing_vit} to accomplish such a challenge, as YOLO models have been validated with good performance deployed on embedded and edge devices \cite{Wenxuan_SpaceOps2025}. Although YOLO architectures have improved in different YOLO versions, small object detection remains challenges. For reducing computational costs, high-resolution images are often resized, while downsampling operations within the network may further weaken the limited features of small objects \cite{liu2020ipg}. Another possible effective method to use high input resolutions to reduce information loss, but this comes at increased computational cost and longer inference time \cite{akyon2022slicing}.

Recent collaborative approach with the use of multiple viewpoints from adjacent satellites provides the way of further improving the detection precision \cite{ai_driven_collab, qu2026collaborative}; however, the essential discussion which are indispensable for viewpoint transmission among a satellite cluster has not been elaborated with full analysis. One concern is whether there is enough capacity for Inter-satellite links (ISLs) to facilitate the transfer of multiple high-resolution pictures. With an increasing  number of support views, communication time may become higher especially when the number of support satellites exceeds the onboard concurrent reception capacity. This may affect the timeliness of the observation results and reduce their value for time-sensitive detection. Therefore, it is necessary to evaluate whether the detection improvements gained from additional support views can justify the associated communication costs.

The main contributions are summarized as:
\begin{itemize}
    \item We design a synthetic SOD dataset (SODv3)\footnote{\url{https://github.com/AEL-Lab/SODv3}} with high-resolution images of small space objects. The dataset includes synchronized multiview observations and metadata describing both observers and targets. It supports single-view and collaborative multiview SOD tasks.
    \item We examine how high-resolution inputs and multiview collaboration affect small SOD. Different YOLO models and fusion strategies are evaluated across varying input resolutions and numbers of support views to compare detection accuracy and inference efficiency. We also assess whether the detection improvements from additional views justify the increased processing and communication costs.
    \item We develop a framework for evaluating communication and processing latency in collaborative SOD scenario, considering ISLs image transmission, limited concurrent reception, and onboard multiview inference. We then examine how ISLs bandwidth, the number of support views, and different fusion strategies affect detection performance and overall latency. This analysis provides insights into the benefits and limitations of collaborative edge AI under different communication constraints.
\end{itemize}

The rest of the paper is structured as follows: Section II discusses the related work; Section III discusses the proposed methodology; Section IV presents the experimental results; and Section V concludes the paper and outlines the future work. 

\section{Related Work}

Detecting small space objects remains challenging due to their limited size, weak visual features, and variations in illumination and background conditions. Previous studies have examined the difficulty of small object detection with limited spatial information \cite{liu2021survey}. Several datasets have been developed for SOD under different observation conditions. The SpotGEO dataset \cite{chen2021spot} contains real observations of geostationary objects captured by ground-based telescopes. These objects often appear as point-like sources or short streaks rather than complete spacecraft structures. The dataset provides image sequences for detecting objects using visual information and motion characteristics. In comparison, the SPARK dataset \cite{rathinam2022spark} contains synthetic spacecraft and debris images with annotations for object detection and perception tasks. 

High-resolution processing and collaborative multiview detection are two possible approaches to improving small SOD performance. For example, Slicing Aided Hyper Inference (SAHI) \cite{akyon2022slicing} divides high-resolution images into smaller regions for detection and then merges the results. This helps preserve visual details that might be lost when resizing the entire image. Collaborative multiview detection instead combines observations of the same target from different satellite viewpoints. These observations may provide additional information when a single view is insufficient for accurate detection. Previous work on collaborative SOD \cite{qu2026collaborative} has explored fusion-based methods that combine observations from multiple satellites to improve detection performance.
\begin{figure*}[h]
    \centering
    \vspace*{0.22in}
    \includegraphics[width=0.9\textwidth,
     trim=1cm 1cm 1cm 1cm
     clip]{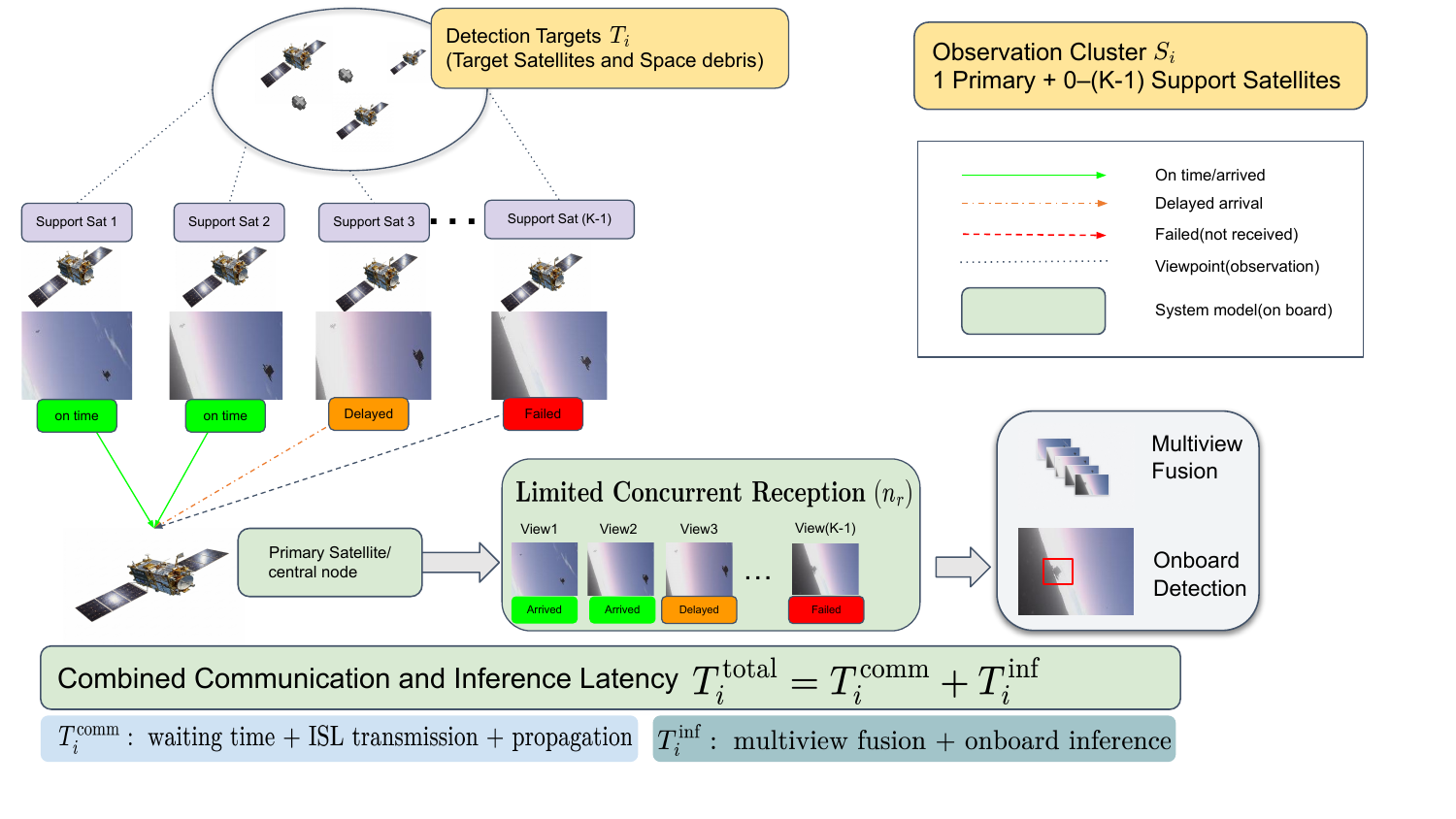}
    \caption{Overview of the proposed cluster-based
    collaborative edge AI framework for multiview
    SOD.}
    \label{fig:system_overview}
    \vspace{-10pt}
\end{figure*}
Both approaches introduce additional resource requirements. High-resolution images increase the amount of data to process and transfer, while multiview detection requires support satellites to send their observations to the primary satellite. The resulting communication and processing costs may affect the time needed for collaborative detection.

ISLs allow data transfer between satellites without the need for data to pass via ground stations (GSs). Recent developments in free-space optical (FSO)-based ISLs have led to high-speed data transfer between Low Earth orbit (LEO) satellites. ISLs have also been explored for collaborative satellite tasks, including task offloading and resource allocation \cite{tong2023inter}. In collaborative SOD, short messages exchanged over ISLs have been used to assist viewpoint selection and reduce unnecessary information transmission \cite{ai_driven_collab}. However, ISL communication still faces practical limitations. Research on dynamic laser ISL routing has examined how link establishment delays and changing network conditions affect communication latency and link availability \cite{bhattacharjee2024demand}. These factors may prevent some support observations from reaching the primary satellite in time for detection.

Previous studies have considered different aspects of satellite cooperation and ISLs communication, but less attention has been paid to the cost of transmitting multiple high-resolution observations for collaborative SOD. This also raises the question of how to design a communication strategy that makes effective use of additional support views. Such a strategy needs to consider whether the potential detection benefits of these views outweigh the transmission delays and communication costs involved.

These challenges motivate us to construct more comprehensive multiview SOD observation scenarios with varying numbers of support views. Based on these scenarios, we investigate detection performance under different processing strategies and analyze communication and inference latency under different ISL capacity and reception constraints.





\section{Proposed Methodology}

\subsection{Multiview Observation Scenario Formulation}

\subsubsection{Observation Cluster}
We consider a cluster-based collaborative edge AI system for SOD, as shown in Fig.~\ref{fig:system_overview}. Multiple observer satellites capture the same target scene from different viewpoints. Support satellites send their observations to the cluster lead through ISLs, where multiview fusion and onboard inference are performed. Let $\mathcal{C} = \{C_1, C_2, \ldots, C_{N_c}\}$ represent the set of observation scenarios, where $N_c$ is the total number of scenarios. We also define that each observation scenario consists of a cluster of observer satellites and a set of external detection targets. So the (i)-th observation scenario is represented as
\begin{equation}
    C_i = (\mathcal{S}_i, \mathcal{T}_i),
    \label{eq:observation_scenario}
\end{equation}
where $\mathcal{S}_i$ denotes the cluster of observer satellites and $\mathcal{T}_i$ denotes the set of external detection targets.

Each observer satellite cluster contains $K$ observer satellites, where $K$ may vary across observation scenarios. The observer satellite cluster in the $i$-th scenario is defined as
\begin{equation}
    \mathcal{S}_i =
    \{s_{i,1}, s_{i,2}, \ldots, s_{i,K}\},
    \label{eq:observer_cluster}
\end{equation}

where $s_{i,j}$ denotes the $j$-th observer satellite in the $i$-th cluster. Within each cluster, one satellite, denoted by $s_{i,p}$, is designated as the primary observer, while the remaining satellites form the support observer set, expressed as
\begin{equation}
    \mathcal{S}_i^{\mathrm{sup}} =
    \mathcal{S}_i \setminus \{s_{i,p}\}.
    \label{eq:support_observers}
\end{equation}

Next, we define the detection target set $\mathcal{T}_i$ for each observation scenario. The detection targets are required to include both external satellites and space debris in the $i$-th scenario, represented as

\begin{equation}
    \mathcal{T}_i =
    \mathcal{T}_i^{\mathrm{sat}} \cup \mathcal{T}_i^{\mathrm{deb}},
    \label{eq:detection_targets}
\end{equation}

where $\mathcal{T}_i^{\mathrm{sat}}$ and $\mathcal{T}_i^{\mathrm{deb}}$ denote the satellite and debris target sets, respectively. These targets
are independent of the observer platforms and are observed by the observer satellites. 

To ensure valid multiview observations, each observer must capture the required satellite and debris. Let $\mathcal{T}_i^{\mathrm{req}} \subseteq \mathcal{T}_i$
denote the set of required targets in the $i$-th observation scenario. The observation constraints are represented as $\mathcal{T}_i^{\mathrm{req}} \subseteq V_{i,j}$ for all $s_{i,j} \in \mathcal{S}_i$, with $\mathcal{T}_i^{\mathrm{req}} \cap \mathcal{T}_i^{\mathrm{sat}} \neq \varnothing$ and $\mathcal{T}_i^{\mathrm{req}} \cap
\mathcal{T}_i^{\mathrm{deb}} \neq \varnothing$,
where $V_{i,j}$ denotes the set of targets satisfying the visibility and annotation conditions in the observation of satellite $s_{i,j}$. These constraints ensure that the required satellite and debris targets are observed across all viewpoints.

\subsubsection{Multiview Observations}
Based on the defined observation scenarios, each observer satellite captures an image from its own viewpoint. The multiview observation set associated with the $i$-th observation scenario is represented as
\begin{equation}
    \mathcal{X}_i =
    \{X_{i,1}, X_{i,2}, \ldots, X_{i,K}\},
    \label{eq:multiview_observations}
\end{equation}
where $X_{i,j}$ denotes the image captured by the $j$-th observer satellite. The primary view corresponding to the primary observer is denoted by $X_{i,p}$, while the remaining observation views form the support-view set, represented as
\begin{equation}
    \mathcal{X}_i^{\mathrm{sup}} =
    \{X_{i,j} \mid j \neq p,\ 1 \leq j \leq K\}.
    \label{eq:support_views}
\end{equation}

For increasing diversity in the images generated, the observation scenarios are formed by introducing changes in the distance from the target, observer positions, and camera orientations. Changes in the observation distance affect the apparent size of the targets. Other changes like spatial shifts and rotations of the camera help to generate different views of the images and reduce centering of the target. In addition, different satellite models, debris variations, Earth and space backgrounds, and simulated glare effects contribute to the visual diversity of the dataset.

To provide a complete description of each observation scenario, the dataset also includes annotations specifying the target categories and bounding-box coordinates in each image. In addition, metadata files are recorded to provide detailed target information, including image positions, observation distances, approximate pixel sizes, and geographic coordinates. The corresponding annotation and metadata sets for the $i$-th observation scenario are represented as
\begin{equation}
    \mathcal{Y}_i =
    \{Y_{i,1}, Y_{i,2}, \ldots, Y_{i,K}\},
    \label{eq:annotation_set}
\end{equation}
\begin{equation}
    \mathcal{M}_i =
    \{M_{i,1}, M_{i,2}, \ldots, M_{i,K}\},
    \label{eq:metadata_set}
\end{equation}
where $Y_{i,j}$ and $M_{i,j}$ denote the annotations and metadata associated with the $j$-th observation, respectively.

The complete multiview observation record in the $i$-th observation scenario is therefore defined as
\begin{equation}
    D_i =
    \{(X_{i,j}, Y_{i,j}, M_{i,j})\}_{j=1}^{K}.
    \label{eq:multiview_record}
\end{equation}
All views belonging to the same observation scenario share a common cluster identifier, preserving the association between the primary and support observations.

\subsection{Small Object Detection and Multiview Processing}
The developed multiview observation scenarios can support different approaches to SOD. First, each individual image will be able to act as a separate single view for detection. This will allow detecting the objects based on the individual view without involving information of other observer satellites. The use of single view detection will also allow comparing this way of detection with more complicated approaches.


The multiview observation scenarios developed support different fusion strategies, including image-level fusion, feature-level fusion, and attention-based fusion. These approaches integrate information from multiple viewpoints at different stages of the detection process.

\textit{Image-Level Fusion}
The primary observation and selected support observations are concatenated along the channel dimension. The fused image input $Z_i^{\mathrm{img}}$ is represented as 
\begin{equation}
   Z_i^{\mathrm{img}} =
    \operatorname{Concat}_{\mathrm{ch}}
    \left(
        X_{i,p}, \mathcal{X}_i^{\mathrm{sup}}
    \right).
    \label{eq:image_level_fusion}  
\end{equation}
where $\operatorname{Concat}_{\mathrm{ch}}$ represents channel-wise concatenation of the views. The combined input is then processed by the detection model to extract features and generate results.

\textit{Feature-Level Fusion:}
Each observation is passed through a shared-weight backbone to extract feature maps at the P3, P4, and P5 scales. At each scale $l$, the features from the
primary and support views are concatenated along the channel dimension and fused as

\begin{equation}
Z_{i,l}^{\mathrm{feat}} =
\phi_l\left(
\operatorname{Concat}_{\mathrm{ch}}
\left(
F_{i,p}^{(l)},
\bar{F}_{i,\mathrm{sup}}^{(l)}
\right)
\right),
\label{eq:feature_fusion}
\end{equation}

where $F_{i,j}^{(l)}$ represents the features from available support
observations are averaged at each scale $l$, and $\phi_l$ consists of a $1\times1$ convolution, batch normalization, and SiLU activation.

\textit{Attention-Based Fusion:}
Features are extracted from each observation using the same shared-weight backbone. The primary view features are used as queries, while the support view features are used as keys and values. The fused features are given by

\begin{equation}
Z_{i,l}^{\mathrm{att}} =
\operatorname{Attention}
\left(Q_{i,p}^{(l)},
K_{i,\mathrm{sup}}^{(l)},
V_{i,\mathrm{sup}}^{(l)}\right),
\label{eq:attention_fusion}
\end{equation}

where $Q_{i,p}^{(l)}$ denotes the query features from the primary view, and
$K_{i,\mathrm{sup}}^{(l)}$ and $V_{i,\mathrm{sup}}^{(l)}$ denote the key and value
features from the support views.


\subsection{Communication Cost Model}
To achieve collaborative multiview detection, an inter-satellite communication model is established to characterize the transmission of support observations to the primary satellite through ISLs. We assume that the primary satellite can simultaneously receive up to $n_r$ support views transmissions, each operating at an effective transmission rate $B$. Once the support satellites have captured their images, they are assumed to be prepared to send them to the primary satellite simultaneously. Let $w_{i,j}$ denote the waiting time of the $j$-th support views. When $K-1 \leq n_r$, all support images can be transmitted simultaneously without
waiting, resulting in $w_{i,j}=0$. Otherwise, the remaining images must wait for receiving capacity to become available. The waiting time for each image depends on the transmission durations of the images assigned earlier.

The communication delay depends on the time required for the last support observation to arrive at the primary satellite, represented as
\begin{equation}
    T_i^{\mathrm{comm}} =
    \max_{\substack{1 \leq j \leq K \\ j \neq p}}
    \left(
        w_{i,j}
        + \frac{\ell_{i,j}}{B}
        + \frac{d_{i,j}}{c}
    \right),
    \label{eq:communication_delay}
\end{equation}
where $\ell_{i,j}$ denotes the image payload size in bits, $d_{i,j}$ is the distance between support satellite $s_{i,j}$ and primary satellite $s_{i,p}$, and $c$ is the speed of light. For $K=1$, no support transmission is required,
and $T_i^{\mathrm{comm}}=0$. The communication payloads are estimated using the real sizes of the compressed JPEG images. The model includes transmission, propagation, and waiting delays, but it does not consider any other forms of communication delays and image processing delays. This allows us to evaluate how the number of support views, transmission rates, and reception capacity affect the communication delay of collaborative multiview detection.

\section{Experimental Results}
\subsection{Experimental Setup}
The experiments are conducted using the constructed SODv3 synthetic SOD dataset generated by Unity. The dataset contains 155,663 images organized into 32,538 synchronized observation groups, with an original image resolution of 3840 × 2160 pixels. Each group consists of one primary observation and up to eight support observations. In addition, the experiments evaluate small object detection at varying input image sizes, compare different multiview fusion strategies and the communication and inference latency for collaborative detection. The computation processing are performed using an NVIDIA H100 GPU with 80 GB of memory. Detection performance is primarily evaluated using mAP@0.5, mAP@0.5:0.95, precision, and recall, while inference time and transmission time are considered
to assess processing efficiency and communication cost. For the communication experiments, the number of support is up to eight, while the effective
ISLs transmission rate $B$ varies from 1\% to 100\% of the nominal rate. 

\subsection{Image Resolution on Detection Performance}

\begin{table}[h]
\centering
\caption{Model (YOLOv11-S) Inference Time at Different Input Resolutions}
\label{tab:inference_time}
\begin{tabular}{lc}
\toprule
Input Resolution & Inference Time (ms) \\
\midrule
640  & 3.06 \\
1280 & 3.40 \\
1920 & 4.12 \\
3840 & 13.61 \\
\bottomrule
\end{tabular}
\end{table}

\begin{figure}[t]
    \centering
    \makebox[\linewidth][c]{%
        \includegraphics[
            width=0.9\linewidth,
            trim=1cm 1cm 1cm 1cm,
            clip
        ]{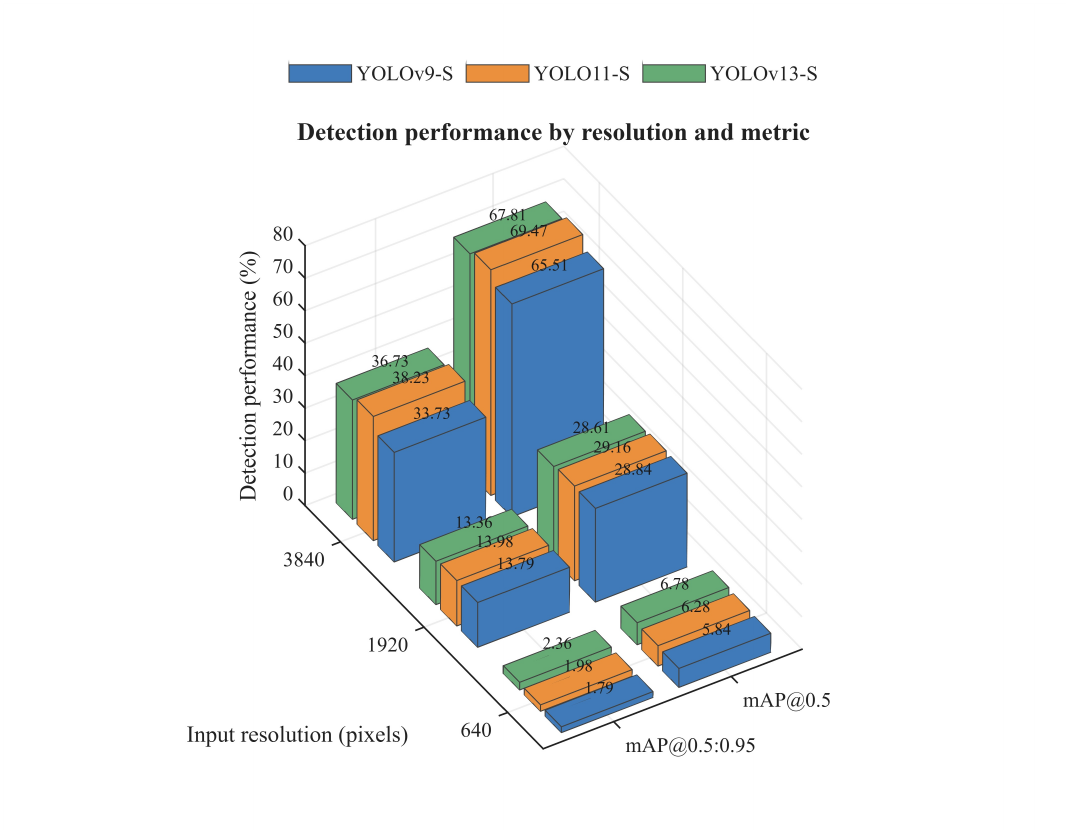}%
    }
    \caption{Detection performance at different input
    resolutions.}
    \label{fig:resolution_comparison}
\end{figure}

We first evaluate single-view SOD using the observations $X_{i,j}$ defined in Section III-A. Each image is processed
independently, without incorporating views from other satellites in the same cluster $S_i$. Fig.~\ref{fig:resolution_comparison} shows the detection performance of YOLOv9-S, YOLOv11-S, and YOLOv13-S at different input resolutions. At the original resolution of $3840 \times 2160$ pixels, YOLO11-S performs best among the three models, achieving an mAP50 of 69.47\% and an mAP50--95 of 38.23\%. Its mAP50 is 3.96 and 1.66 percentage points higher than those of YOLOv9-S and YOLOv13-S, respectively.

Detection performance decreases considerably as the input resolution is reduced. For YOLOv11-S, mAP50 drops
from 69.47\% at 3840 pixels to 29.16\% at 1920 pixels and 6.28\% at 640 pixels.

The decline in detection performance at lower resolutions suggests that downsampling removes important visual
details needed to detect small space objects. For YOLOv11-S, increasing the input resolution from 1920 to 3840 pixels
improves mAP50 by 138.2\%. However, the inference time also increases from 4.12 ms to 13.61 ms, approximately
3.30 times as long, as shown in Table~\ref{tab:inference_time}. This trade-off between detection accuracy and inference latency is particularly relevant when deploying SOD models on resource-constrained satellite platforms.

\begin{figure}[t]
    \centering
    \includegraphics[width=0.8\columnwidth]
    {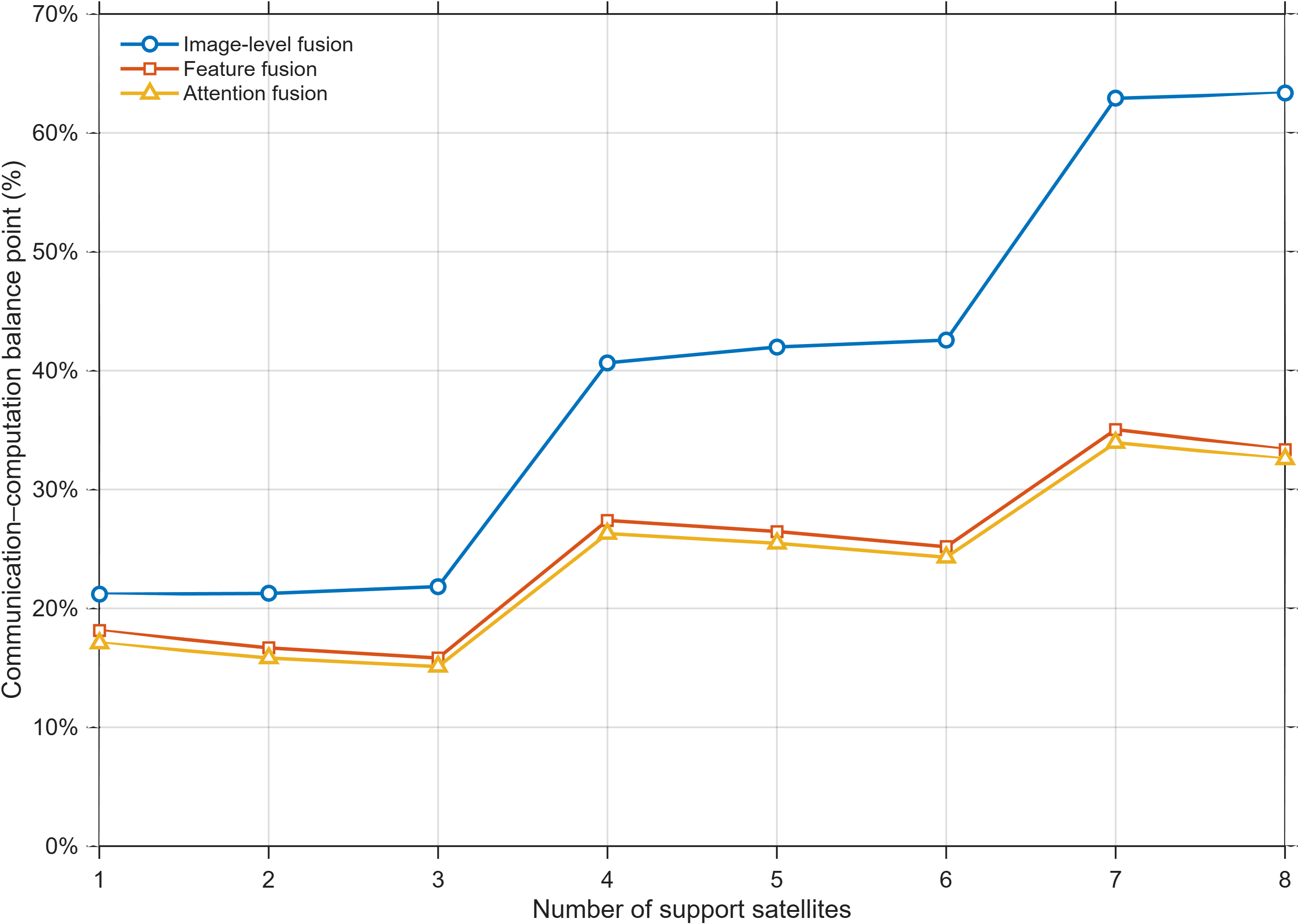}
    \caption{Communication--computation balance points of three
    multiview fusion strategies.}
    \label{fig:communication_computation_balance}
\end{figure}

\begin{figure*}[t]
    \centering
    \includegraphics[
        width=1\textwidth,
    ]{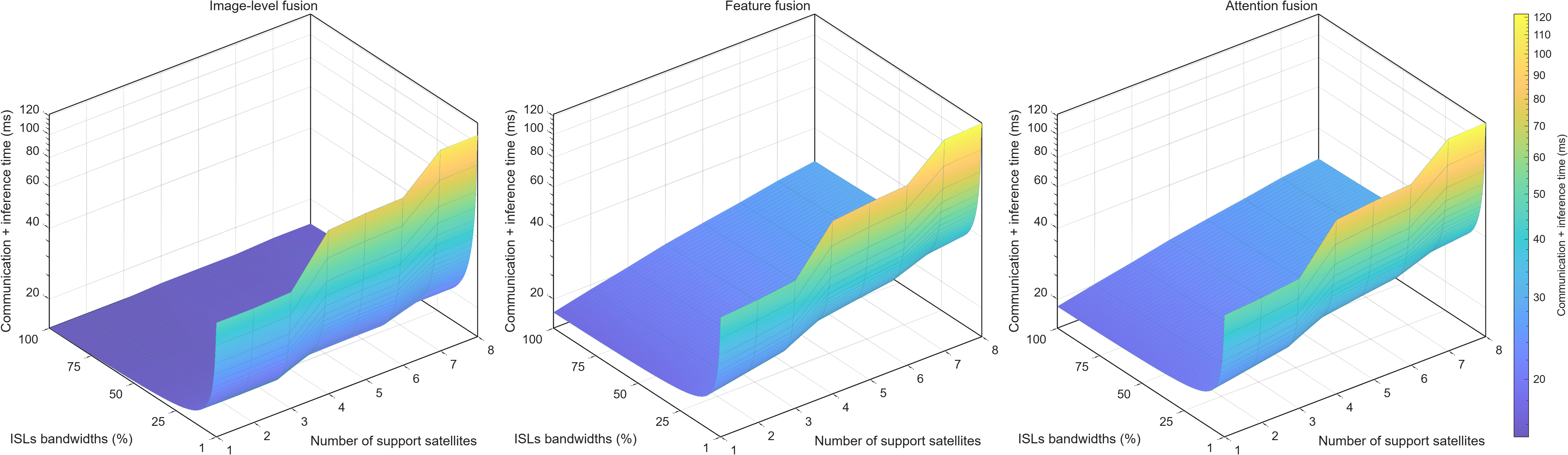}
    \caption{Combined Communication and Inference Latency of Three Multiview Fusion Strategies}
    \label{fig:three_methods_communication_cost}
    \vspace{-15pt}
\end{figure*}

\begin{table}[t]
\centering
\caption{Detection Performance of Multiview Fusion
Strategies with Different View Counts (\%)}
\label{tab:multiview_fusion}
\setlength{\tabcolsep}{3pt}
\begin{tabular}{lcccc}
\toprule
& \multicolumn{2}{c}{All Views}
& \multicolumn{2}{c}{$K \leq 3$} \\
\cmidrule(lr){2-3}
\cmidrule(lr){4-5}
Fusion Strategy
& mAP50 & mAP50--95
& mAP50 & mAP50--95 \\
\midrule
Image-Level
& 49.57 & 18.66
& \textbf{53.64} & \textbf{21.84} \\
Feature-Level
& \textbf{52.74} & 22.93
& 50.76 & 20.86 \\
Attention-Based
& 52.45 & \textbf{23.65}
& 50.96 & 20.20 \\
\bottomrule
\end{tabular}
\end{table}

Table~\ref{tab:multiview_fusion} compares the detection performance of three multiview fusion strategies using YOLOv11-S as the base detector. None of the three strategies outperformed the single-view baseline, suggesting that additional support views do not always improve detection accuracy. For image-level fusion, restricting the evaluation to clusters with one to three views increased mAP50 from 49.57\% to 53.64\%. This suggests that using fewer observations may provide better detection results without requiring the transmission of all available support views.

From a communication perspective, selecting informative support views could reduce unnecessary image transmissions and the associated ISL communication latency. Another possibility is to transmit compressed feature maps instead of full images to reduce the amount of data transferred. These approaches could help reduce communication costs in collaborative SOD, although their effects on detection performance and processing efficiency require further investigation.

\subsection{Inter-Satellite Communication and Processing Latency}

Fig.~\ref{fig:three_methods_communication_cost} compares the combined communication and inference latency of three multiview fusion strategies under varying numbers of support satellites and ISL bandwidths. Increasing ISL bandwidth reduces overall latency for all three strategies, particularly at lower bandwidths where image transmission accounts for much of the total delay. At higher bandwidths, inference time becomes more dominant, limiting the benefit of further bandwidth increases.

Image-level fusion achieves the lowest combined latency, with average reductions of 30.13\% and 32.80\% compared with feature-level and attention-based fusion, respectively. At low ISL bandwidths (1\%--10\% of nominal capacity), these reductions are 20.79\% and 22.92\%, increasing to 32.23\% and 34.99\% at higher bandwidths (80\%--100\%). This difference comes from the processing structure. Image-level fusion uses a single detector for the concatenated observations, while the other two strategies extract features from each view separately. As discussed in Section IV-B, image-level fusion also provides better detection accuracy for clusters with up to three views. When all available views are included, feature-level and attention-based fusion achieve higher detection accuracy but require more inference time.

Fig.~\ref{fig:communication_computation_balance} further explore the communication--computation balance point, defined as the per-link data rate at which $T_{\mathrm{comm}}=T_{\mathrm{infer}}$, expressed as a percentage of the nominal per-link ISL capacity of 1~Gbps. Below this point, communication latency is higher than inference latency, while inference dominates above it. With one support view, the balance points are 21.19\%, 18.12\%, and 17.09\% for image-level, feature-level, and attention-based fusion, respectively. With eight support views, they increase to 63.34\%, 33.37\%, and 32.57\%. The larger increase for image-level fusion is related to its nearly constant inference time, while the increasing inference costs of feature-level and attention-based fusion partly offset the effect of additional communication delays on their balance points.

All three fusion strategies show sharp increases in their balance points when the number of support views exceeds a multiple of the concurrent reception capacity ($n_r$). In these cases, an additional transmission round is needed, introducing further waiting delays. Between these increases, the balance points of feature-level and attention-based fusion decrease slightly as inference time grows with the number of views. This pattern shows how reception capacity and fusion processing costs together affect the ISL bandwidth at which inference latency becomes greater than communication latency.

The communication--computation balance points may also provide a basis for adaptive support-view admission. For each fusion strategy and number of support views, the balance point can serve as a reference bandwidth threshold. Comparing the values before and after adding a support view helps determine whether the main latency bottleneck would shift from inference to communication under the available ISL bandwidth. This is particularly relevant when the number of support views exceeds the concurrent reception capacity, as additional transmission rounds introduce waiting delays. The decision to include more views would also depend on their expected detection benefits and the total latency budget. Such an approach could support communication-aware support-view selection in future collaborative SOD systems.

\section{Conclusion}
This paper explored cluster-based collaborative edge AI for small SOD using the SODv3 dataset. Higher input resolutions improved detection accuracy but required longer inference time. Adding more support views did not always improve detection performance, and different fusion strategies showed varying results in detection accuracy and processing efficiency. Communication latency was also affected by ISL bandwidth and concurrent reception capacity, with substantial delays occurring when the reception limit was exceeded. These results suggest that the number of support views and fusion strategy should be selected by considering both detection performance and communication and processing costs.

\ifCLASSOPTIONcaptionsoff
  \newpage
\fi




\bibliographystyle{IEEEtran}
\bibliography{references}




\end{document}